%% file: main.tex
\PassOptionsToPackage{table,dvipsnames}{xcolor}

\documentclass[letterpaper, 10 pt, conference]{ieeeconf}

\IEEEoverridecommandlockouts
\usepackage[utf8]{inputenc}
\usepackage[T1]{fontenc}
\usepackage[scaled=.92]{helvet} %
\usepackage{cite}
\usepackage{amsmath,amssymb}
\usepackage{graphicx}
\usepackage{booktabs}
\usepackage{multirow}
\usepackage{array}
\usepackage{tabularx}
\usepackage{placeins}
\usepackage{pifont}
\usepackage{xcolor}
\usepackage{colortbl}  %
\usepackage[hidelinks,bookmarks=false]{hyperref}
\usepackage{url}
\graphicspath{{figures/}}

\newcommand{\ours}{RYOPO}
\newcommand{\oursrow}[1]{\textbf{#1\,(ours)}}

\DeclareRobustCommand{\modulelabel}[1]{\textup{\textbf{(#1)}}}
\title{\LARGE \bf
RYOPO: Bringing End-to-End Category-Level Object Pose Estimation into Real Time}

\author{Hakjin Lee \qquad Junghoon Seo \qquad Jaehoon Sim%
\thanks{PIT IN Co., Anyang-si, Gyeonggi-do, Republic of Korea.
Email: \{hj, sjh, simjeh\}@pitin-ev.com}}

\begin{document}

\maketitle
\thispagestyle{empty}
\pagestyle{empty}

\input{sections/abstract}
\input{sections/introduction}
\input{sections/related_work}
\input{sections/method}
\input{sections/experiments}
\input{sections/conclusion}

\bibliographystyle{IEEEtran}
\begingroup
\def\BIBdecl{\setlength{\itemsep}{-0.75pt}}
\bibliography{main}
\endgroup

\clearpage
\input{sections/appendices}
\FloatBarrier

\end{document}

%% file: sections/abstract.tex
\begin{abstract}
Category-level object pose estimation predicts the rotation, translation, and metric size
of unseen instances within known categories.
Many accurate RGB-D methods rely on external instance segmentation and crop-based pose
estimation, introducing separate stages and object-dependent processing costs that hinder
real-time inference.
To bring accurate pose estimation into real time, we present \ours{}, an end-to-end trainable
query-based RGB-D set predictor. It jointly detects and segments objects and estimates their
\mbox{9-DoF} poses without explicit CAD-derived shape priors or a separately trained instance segmentor.
Shared image and scene encoding avoids repeated per-object crop encoding.
A query-conditioned geometry pathway associates observed 3D points and RGB features with
object queries and incorporates shared scene context.
Object-centric refinement uses the resulting point descriptors to update an explicit pose state
through pose-conditioned cross-attention and recurrent residual corrections.
On NOCS, \ours{} substantially improves on published RGB-D joint detection and pose estimation results.
It achieves competitive performance compared with two-stage methods
under all-object evaluation on REAL275 and HouseCat6D, while enabling
real-time full-frame pose estimation at $31.8$ FPS on an RTX~A6000.
Project page: \url{https://yopo-series.github.io/RYOPO-project-page/}.
\end{abstract}

%% file: sections/introduction.tex
\section{Introduction}
\label{sec:intro}
\suppressfloats[t]
\input{figures/captions/figure1_runtime}

Category-level object pose estimation predicts poses and metric sizes for unseen instances
within known categories, without test-instance CAD models~\cite{nocs}.
Common RGB-D pipelines use a separately trained instance segmentor to obtain object masks
and extract each object's RGB crop and depth-derived point cloud for a per-object pose
network~\cite{agpose,cleanpose}.
This design adds an external segmentation stage and repeats crop encoding and pose
estimation for each detected instance, increasing inference cost with object count.
The resulting inference latency can limit closed-loop visual-servoing
performance~\cite{corke1996dynamic}.
It can also reduce manipulation throughput by extending task cycles when perception
cannot be overlapped with robot motion~\cite{fu2024activebin}.

Joint methods already combine localization and pose prediction. CenterSnap~\cite{centersnap}
and ShAPO~\cite{shapo} use learned CAD-derived shape representations for RGB-D inference,
whereas YOPO~\cite{yopo} extends a detection transformer with parallel pose heads on
image-derived queries. We aim to extend the set-prediction formulation used by YOPO
to end-to-end RGB-D object pose estimation while achieving real-time inference.
The challenge is to gain accuracy from measured 3D geometry while keeping full-frame
computation within a real-time budget. This favors shared image and scene encoding over
independently encoding a crop for every object.

A straightforward RGB-D extension is to fuse encoded depth and RGB features before query decoding.
Early fusion makes depth available through shared features, but does not by itself specify
how each object's measured 3D geometry should guide pose corrections.
In reference-box-based detection transformers such as DINO~\cite{dinodetr}, queries combine
content features with positional embeddings derived from 2D boxes. These spatial references
support image localization but do not explicitly represent 3D rotation, translation, or metric size.
We therefore investigate object-specific pose corrections guided by measured geometry and
the current pose estimate.

We present \textbf{\ours{}} (\textbf{R}eal-Time \textbf{YOPO}), a query-based RGB-D set-prediction framework for joint
detection, segmentation, and category-level object pose estimation. To the best of our
knowledge, \ours{} is the first end-to-end trainable query-based RGB-D set predictor for
category-level object pose estimation without explicit CAD-derived shape priors.
Built on the DETR-style EdgeCrafter architecture~\cite{liu2026edgecrafter}, \ours{}
learns localization and pose estimation jointly, without a separately trained instance
segmentor or per-object crop encoder.

The key design couples object-associated geometry with an explicit pose state.
A query-conditioned geometry pathway associates measured 3D observations with object queries
and combines them with RGB appearance and shared scene context to update query content
and positional representations. Unlike YOPO's parallel pose heads, object-centric refinement
combines object queries with current pose encodings for cross-attention to object
observations. Queries predict residual pose corrections; updated poses condition
further refinement without test-time optimization.

On NOCS, \ours{} substantially improves on published RGB-D joint detection and pose estimation
results. On REAL275 and HouseCat6D, performance is competitive with two-stage methods
under all-object evaluation. Our deployment achieves $31.8$ FPS for full-frame pose estimation
on an RTX~A6000 (Fig.~\ref{fig:teaser}).

Our main contributions are:
\begin{itemize}
\item An end-to-end trainable RGB-D set predictor for joint detection, segmentation,
and object pose estimation without explicit CAD-derived shape priors or an external
instance segmentor.
\item An object-level depth integration method that associates observed 3D geometry with
DETR-style object queries, and a pose refinement method that combines query features
with the current pose state to attend to object observations and predict residual pose corrections.
\item Substantial gains over RGB-D joint methods on NOCS and competitive all-object
pose performance on REAL275 and HouseCat6D, with real-time pose inference.
\end{itemize}

%% file: figures/captions/figure1_runtime.tex
\begin{figure}[t]
\centering
\includegraphics{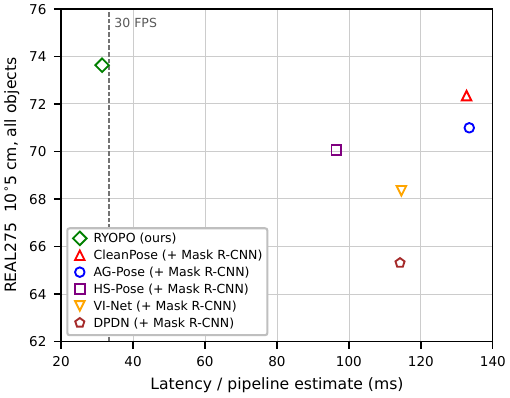}
\caption{REAL275 all-object $10^\circ5$\,cm AP versus mean pose-output latency from Table~\ref{tab:latency} on RTX~A6000.}
\label{fig:teaser}
\label{fig:frontier}
\end{figure}

%% file: sections/related_work.tex
\section{Related Work}

\noindent\textbf{Category-level object pose estimation.}
A central challenge is learning representations that generalize across
intra-category shape variation. NOCS~\cite{nocs} estimates pose and size by aligning
observed points with a shared canonical coordinate space.
SPD~\cite{spd} and GCE-Pose~\cite{gcepose}
deform category shape priors, while
DPDN~\cite{dpdn} combines prior deformation with direct pose regression.
Methods without explicit CAD-derived shape priors learn geometric
representations~\cite{vinet,hspose}, adaptive keypoints~\cite{agpose}, or features
guided by causal learning and knowledge distillation~\cite{cleanpose}.
AG-Pose and CleanPose use instance-level 3D models for NOCS training
supervision, requiring shape data in addition to pose annotations.
\ours{} constructs object-specific representations from predicted boxes and masks
within a joint predictor, without instance 3D models or explicit shape priors.

\noindent\textbf{Joint object detection and pose estimation.}
Joint methods integrate object localization and pose estimation within a shared model.
CenterSnap~\cite{centersnap} and ShAPO~\cite{shapo} jointly estimate RGB-D object poses
and reconstruct shapes using CAD-derived representations. ShAPO also uses test-time optimization.
FSD~\cite{fsd} also trains RGB-D detection and pose estimation end to end using center heatmaps.
It uses a frozen DeepSDF decoder pretrained on ShapeNet as a CAD-derived shape prior
for reconstruction and self-supervised learning.
YOPO~\cite{yopo} instead uses query-based set prediction for RGB-only pose estimation
without explicit shape supervision. Its decoder refines image-derived queries before
parallel heads predict pose.

This query-based formulation builds on DETR~\cite{detr} and its
extensions~\cite{deformabledetr,dinodetr,maskdino}, while the
RT-DETR series~\cite{rtdetr,rtdetrv2,rtdetrv3,rtdetrv4} targets real-time detection.
Spatial encoding~\cite{petr,detr3d,vdetr} and point--voxel cross-attention~\cite{uni3detr}
also guide queries in $3$D detection. For efficient RGB-D pose estimation, we adopt
EdgeCrafter~\cite{liu2026edgecrafter} as our detector foundation.
\ours{} extends this foundation by conditioning attention
to observed object geometry on an evolving 3D pose state.

\noindent\textbf{Object- and scene-level feature fusion.}
RGB and point-cloud features provide complementary appearance and geometry cues.
For pose estimation, object-level fusion combines these cues within supplied object regions.
DenseFusion~\cite{densefusion} uses dense feature fusion and iterative refinement for
instance-level pose estimation; SecondPose~\cite{secondpose} fuses geometric and
pretrained semantic features for category-level prediction.

Scene-level image--point-cloud fusion extends integration to multi-object scenes through
early point/voxel feature augmentation~\cite{mvxnet}, query-mediated integration~\cite{brt,cmt},
or feature exchange~\cite{tokenfusion}. \ours{} combines shared image and scene encoding
with query-associated object observations, enabling object-specific pose refinement
within a joint set predictor.

%% file: sections/method.tex
\section{Method}
\input{figures/captions/figure2_architecture}

\label{sec:problem}
\label{sec:formulation}
Given an RGB image $\mathbf I\in\mathbb R^{H\times W\times3}$, an aligned depth map
$\mathbf D\in\mathbb R_{\geq0}^{H\times W}$, and camera intrinsics $\mathbf K\in\mathbb R^{3\times3}$,
where $H,W$ are the image height and width, \ours{} predicts
objects with category $c$, box $\mathbf b$, mask $\mathbf M$, and \mbox{9-DoF} pose $(\mathbf R,\mathbf t,\mathbf s)$, where
$\mathbf R\in SO(3)$, $\mathbf t\in\mathbb R^3$, and $\mathbf s\in\mathbb R^3_{>0}$ denote rotation, translation,
and metric size along the canonical axes.
Each object query predicts these attributes jointly and selects observations through
its box and mask, without an external instance segmentor or explicit CAD-derived shape priors.

To target accurate real-time inference, \ours{} replaces YOPO's DINO
detector~\cite{yopo,dinodetr} with the DETR-style EdgeCrafter architecture~\cite{liu2026edgecrafter},
retaining its shared RGB backbone, image encoder, and object-query decoder
(unfilled blocks in Fig.~\ref{fig:arch}). As in YOPO, we add parallel
coarse-pose heads alongside the detection and mask heads. For RGB-D input, we use
observed depth to anchor query references and coarse translation estimates.

Our EdgeCrafter-based direct predictors with depth-based query initialization or
early RGB--depth fusion~\cite{mvxnet} (Table~\ref{tab:components}) achieve lower pose accuracy
than state-of-the-art per-object methods~\cite{agpose,cleanpose} under all-object evaluation
on REAL275. This motivates an integrated geometry pathway that connects query-associated
observations, shared scene context, and explicit pose-state refinement.

During set decoding (Fig.~\ref{fig:arch}), \modulelabel{A} forms point descriptors from the current boxes and masks.
Module \modulelabel{B} combines these observations with shared scene context to update queries for the next decoder layer.
After set decoding, \modulelabel{C} keeps the final descriptors fixed while iteratively updating pose.
The image and scene encoders are shared; no per-object crop encoder is added.
We omit the object-query index below and use $d$ for the shared feature width.

\subsection{Query-conditioned geometry \modulelabel{A}}
\label{sec:mpt}
Each query needs object-associated depth observations for feedback and refinement.
Its box defines a sampling lattice, while mask confidence and local depth agreement
reduce contributions from the background and inconsistent surfaces.

\textbf{Point construction.}
Back-projection of $\mathbf D$ with $\mathbf K$ gives a camera-space point map.
The lattice is divided into $G\times G$ cells.
At each candidate location $\mathbf u_i\in[-1,1]^2$ in normalized full-image coordinates, bilinear sampling of predicted
mask probabilities and valid-depth indicators gives $m_i,v_i\in[0,1]$.
Validity-weighted interpolation of the point map gives
$\mathbf p_i\in\mathbb R^3$, with camera-axis depth $d_i=(\mathbf p_i)_z$.
The base weight is $b_i=m_iv_i$.
Within each cell $C$, we initialize the reference depth $\tilde d_C$ as the weighted mean
of its candidate depths, then compute depth-consistency weights $\omega_i$ and pool a 3D point:
\begin{equation}
\begin{aligned}
\tilde d_C &= \frac{\sum_{i\in C}b_i d_i}{\sum_{i\in C}b_i},\\
\omega_i &= b_i\exp\!\left(-\frac{|d_i-\tilde d_C|}{\tau\tilde d_C}\right),\\
\mathbf p_C &= \frac{\sum_{i\in C}\omega_i\,\mathbf p_i}{\sum_{i\in C}\omega_i}.
\end{aligned}
\label{eq:select}
\end{equation}
Here $\tau>0$ is the relative-depth tolerance and $\mathbf p_C\in\mathbb R^3$ the pooled point.
The exponential kernel downweights relative-depth disagreement.
We perform two weighting passes, updating $\tilde d_C$ between them by the same
weighted mean with $b_i$ replaced by $\omega_i$, then pool points using the final weights.

\textbf{Object point descriptors.}
Using the final depth-consistency weights in Eq.~\eqref{eq:select}, we pool
$m_i$, $v_i$, and $\mathbf u_i$ using the same weighted mean as for $\mathbf p_i$,
giving $m_C,v_C\in[0,1]$ and $\mathbf u_C\in[-1,1]^2$.
We sample the image feature $\mathbf f_C\in\mathbb R^d$ at $\mathbf u_C$ and define
point support as $a_C=m_Cv_C$. Let $\boldsymbol\mu\in\mathbb R^3$ and $\rho\geq0$ denote the centroid and
root-mean-square radius of the pooled points, weighted by $a_C$. We form $\mathbf z_C\in\mathbb R^{d+16}$ as
\begin{equation}
\mathbf z_C=[\,(\mathbf p_C-\boldsymbol\mu)/\rho,\;\mathbf p_C-\boldsymbol\mu,\;\mathbf p_C,\;\mathbf f_C,\;\boldsymbol\eta_C\,],
\label{eq:token}
\end{equation}
where $\boldsymbol\eta_C=[m_C,v_C,\mathbf r_C,\mathbf u_C]\in\mathbb R^7$ collects support and viewing information,
and brackets denote concatenation.
Here $\mathbf r_C=\mathbf p_C/(\mathbf p_C)_z\in\mathbb R^3$ is the depth-normalized camera ray.
Normalized offsets encode local shape; metric offsets and camera-space XYZ retain size
and position; image features provide appearance. The $G^2$ descriptors
$\mathcal Z=\{\mathbf z_C\}$ provide inputs to modules \modulelabel{B} and \modulelabel{C}.

\subsection{Geometry feedback \modulelabel{B}}
\label{sec:geometry}
Geometry feedback lets 3D observations guide object-query updates during set decoding,
before final pose refinement.
The parallel branches in Fig.~\ref{fig:arch} update query content $\mathbf q\in\mathbb R^d$
and positional representation $\boldsymbol\pi\in\mathbb R^d$ for the next decoder layer.
The positional representation retains the decoder's 2D reference-box embedding
and is augmented by geometry feedback.

\textbf{Shared scene context.}
The Scene Context Encoder back-projects and voxelizes the depth map, then applies sparse
3D convolutions~\cite{submanifold} to obtain a scene feature bank $\mathcal S\subset\mathbb R^d$.
Computed once per image, this bank supplies context beyond each object's observations.
Scene cross-attention uses embeddings of query boxes and voxel positions,
producing the scene-attention output $\mathbf u_s\in\mathbb R^d$.

\textbf{Object cross-attention.}
Let $\nu_C=f_v(a_C)$ denote soft point validity, where $f_v$ is a fixed sigmoid mapping.
Its logarithm gives the attention bias $\beta_C=\log\nu_C$.
We write $\operatorname{CA}_{\boldsymbol\beta}$ for multi-head cross-attention with
$\beta_C$ added to each point's attention logit before softmax.
To combine the descriptors with query features, we first embed each
point descriptor using \rule{0pt}{14pt}$E_o:\mathbb R^{d+16}\to\mathbb R^d$.
The object output
$\mathbf u_o\in\mathbb R^d$ and feedback updates are
\begin{equation}
\begin{aligned}
\mathbf u_o &= \operatorname{LN}\!\left(\mathbf q+
\operatorname{CA}_{\boldsymbol\beta}(\mathbf q,E_o(\mathcal Z))\right),\\
\mathbf q^+ &= \mathbf q+g(\mathcal Z)\,F_c(\mathbf u_o)+F_s(\mathbf u_s),\\
\boldsymbol\pi^+ &= \boldsymbol\pi+g(\mathcal Z)\,F_\pi(\mathbf u_o),
\end{aligned}
\label{eq:feedback}
\end{equation}
where $F_c,F_\pi,F_s:\mathbb R^d\to\mathbb R^d$ are learned projections with bounded output scaling.
$\operatorname{LN}$ is layer normalization and superscript $+$ denotes the post-feedback state.
$F_c$ and $F_\pi$ map $\mathbf u_o$ to query content and positional updates, respectively;
$F_s$ maps $\mathbf u_s$ to a content update.
The separate object-level support gate $g(\mathcal Z)=\max_C\nu_C\in[0,1]$ suppresses object feedback
when all sampled points have low validity.
Scene attention similarly includes a query residual and layer normalization.
Both branches feed the next set-decoder layer.

\subsection{Object-centric pose refinement \modulelabel{C}}
\label{sec:decoder}
Unlike YOPO's parallel pose heads on image-derived decoder queries~\cite{yopo},
module \modulelabel{C} maintains a pose tuple $P=(\mathbf R,\mathbf t,\mathbf s)$.
The final query and point descriptors guide pose-conditioned cross-attention and residual updates
(Fig.~\ref{fig:arch}).
Conditioning attention on the current pose allows the model to interpret object observations
relative to its present estimate. Each layer predicts a correction, and the updated pose
conditions the next refinement step.

\textbf{Pose-conditioned cross-attention.}
\label{sec:refinement}
The final object query initializes the latent state $\mathbf h^{(0)}\in\mathbb R^d$.
To encode observation support and spatial extent, we use
$\mathbf s_{\mathrm{obs}}=[G^{-2}\sum_C\nu_C,\boldsymbol\sigma_p]\in\mathbb R^4$,
where $\boldsymbol\sigma_p\in\mathbb R^3$ contains the axis-wise standard deviations of
pooled points about $\boldsymbol\mu$, with normalized weights $a_C/\sum_Da_D$.
The pose embedding $\phi$ maps a 16D concatenation of translation, log size,
a 6D rotation representation~\cite{rot6d}, and $\mathbf s_{\mathrm{obs}}$ to $\mathbb R^d$.
The conditioning pose $\bar P^{(m-1)}$ is the coarse prediction in the first layer
and the preceding layer's updated pose thereafter.
With point validities $\boldsymbol\nu=(\nu_C)_C\in[0,1]^{G^2}$ and pointwise embedding
$E_r:\mathbb R^{d+16}\to\mathbb R^d$, layer $m$ computes
\begin{equation}
\begin{aligned}
\tilde{\mathbf q}^{(m)} &= \mathbf h^{(m-1)}+\phi(\bar P^{(m-1)},\mathbf s_{\mathrm{obs}}),\\
\mathbf h^{(m)} &= \mathcal A_m(\tilde{\mathbf q}^{(m)},E_r(\mathcal Z);\boldsymbol\nu),\\
(\boldsymbol\delta_t^{(m)},\boldsymbol\delta_s^{(m)},\boldsymbol\delta_r^{(m)}) &= H_m(\mathbf h^{(m)}).
\end{aligned}
\label{eq:psr}
\end{equation}
The pose encoding augments the attention query, while the final point descriptors
provide the keys and values.
The block $\mathcal A_m$ weights embedded descriptors by validity and uses the same
pre-softmax log-validity bias as \modulelabel{B}. Residual connections, normalization, and a feed-forward layer yield the updated query.
The head $H_m:\mathbb R^d\to\mathbb R^9$ returns bounded translation, log-size, and axis-angle increments.

\textbf{Residual pose updates.}
At each layer, rotation is updated by composition,
camera-frame translation by size-scaled increments, and size in log space:
\begin{equation}
\begin{aligned}
\mathbf R^{(m)} &= \exp([\boldsymbol\delta_r^{(m)}]_\times)\,\mathbf R^{(m-1)},\\
\mathbf t^{(m)} &= \mathbf t^{(m-1)}+\boldsymbol\delta_t^{(m)}\odot\mathbf s^{(m-1)},\\
\log\mathbf s^{(m)} &= \log\mathbf s^{(m-1)}+\boldsymbol\delta_s^{(m)}.
\end{aligned}
\label{eq:compose}
\end{equation}
Here the mapping from an axis-angle vector to its skew-symmetric matrix is
$[\cdot]_\times:\mathbb R^3\to\mathbb R^{3\times3}$;
$\exp$ is the matrix exponential, and $\odot$ denotes element-wise multiplication.
These updates preserve valid rotations and positive sizes.
The updated pose is re-encoded for the
next refinement layer. These are learned forward-pass updates, with no inference-time
gradient optimization.

\subsection{Training}
\label{sec:supervision}
We jointly train detection, segmentation, and pose prediction using pose-aware
one-to-one Hungarian assignment~\cite{yopo}. For each supervised output set, the
same assignment pairs a query's class, box, mask, and pose with one annotated object;
unmatched queries receive negative class supervision. A weighted sum of detection/mask,
pose, and auxiliary geometry losses trains the joint predictor, with pose supervision
on coarse predictions and refinement outputs. Pose losses supervise center, depth, extent,
and rotation and include geodesic rotation and symmetry-aware 3D/projected cuboid terms.
To encourage geometrically consistent object representations, auxiliary losses supervise
camera-space keypoints and their canonical coordinates.
Learned auxiliary class-conditioned queries aggregate observed points into
camera-space keypoints $\mathbf x_k\in\mathbb R^3$; their features predict canonical coordinates
$\mathbf y_k\in\mathbb R^3$, with $k$ indexing keypoints.
These queries are learned parameters, not CAD-derived templates.
Annotated pose and size
$(\mathbf R^\star,\mathbf t^\star,\mathbf s^\star)$ define the canonical-keypoint target
\mbox{$\mathbf y_k^\star=(\mathbf R^\star)^\top(\mathbf x_k-\mathbf t^\star)/\|\mathbf s^\star\|_2$};
coverage and diversity regularizers distribute keypoints over the observed support.
Contrastive denoising~\cite{dinodetr} supplies additional class/pose supervision through
attention-isolated queries during training only.

%% file: figures/captions/figure2_architecture.tex
\begin{figure*}[t]
  \vspace*{4pt}
  \centering
  \raisebox{-0.0pt}[231.72389pt][0.0pt]{%
  \makebox[505.89813pt][l]{\includegraphics{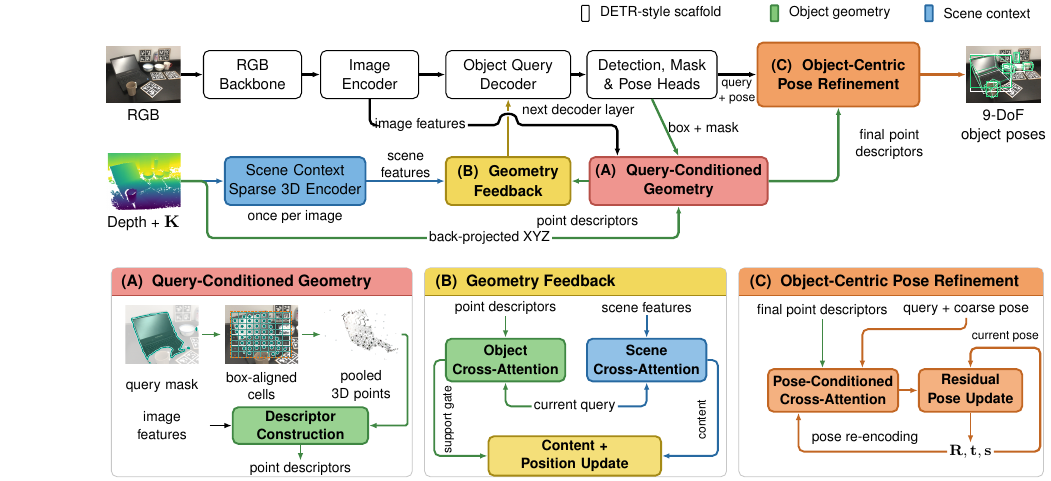}}}
  \caption{End-to-end RGB-D set prediction in \ours{}. \modulelabel{A} query-conditioned geometry and \modulelabel{B} geometry feedback support \modulelabel{C} object-centric pose refinement. Unfilled blocks form the DETR-style scaffold. Point colors come from RGB; marker size and opacity encode support, and hollow markers indicate low support. Output visualization: predicted cuboids (green), annotations (white).}
  \label{fig:arch}
  \vspace*{-4pt}
\end{figure*}

%% file: sections/experiments.tex
\section{Experiments}
\label{sec:results}
\subsection{Experimental setup}
\label{sec:protocol}
\label{sec:implementation}
\noindent\textbf{Datasets.}
We evaluate joint RGB-D set prediction on NOCS~\cite{nocs} and
HouseCat6D~\cite{housecat6d}. NOCS covers six categories, with synthetic and real-world RGB-D data
evaluated on CAMERA25 and REAL275, respectively;
HouseCat6D contains ten household categories.
Evaluation uses $2{,}754$ REAL275 and $2{,}929$ HouseCat6D frames, and the
$19{,}220$ CAMERA25 frames in the released AG-Pose/CleanPose prediction records.

\input{tables/nocs_joint}

\noindent\textbf{Evaluation metrics.}
We follow CleanPose~\cite{cleanpose}, reporting mean average precision (AP) across categories.
\textbf{3D IoU:} We use CATRE~\cite{catre} IoU$_{75}$ for NOCS and independent
IoU$_{25}$ and IoU$_{50}$ for HouseCat6D.
Tables~\ref{tab:nocs} and~\ref{tab:coverage} additionally report independent IoU$_{50}$.
\textbf{\boldmath$n^\circ m$\,cm:} A pose is correct when its symmetry-aware
rotation error is within $n^\circ$ and its translation error within $m$\,cm.
\textbf{All-object evaluation:} We disable CleanPose's $0.1$ IoU prefilter,
retaining every annotation and all submitted predictions. IoU-matched AP applies
this prefilter to both annotations and predictions.
Thus, objects missed by a pipeline or rejected during preprocessing remain in
the evaluation rather than disappearing from the pose-scoring population.

\noindent\textbf{Implementation and training.}
We use the DETR-style EdgeCrafter~\cite{liu2026edgecrafter} architecture with its
pretrained RGB backbone.
The model uses $640\times640$ inputs,
$150$ object queries, four set-decoder layers, and three pose-refinement layers.
Each query uses $64$ point descriptors from an $8\times8$ grid; scene encoding uses $5$\,mm voxels.
We train with AdamW~\cite{adamw}, an effective batch size of $72$, mixed precision, and weight decay $10^{-4}$.
NOCS training combines real images with $38{,}900$ CAMERA images redrawn from
the full training pool each epoch.
NOCS and HouseCat6D use $12$ and $18$ epochs, with learning-rate decays at $9$/$11$ and
$14$/$17$, respectively. At a reference batch size of $16$, the base learning rate is
$5\times10^{-4}$. We use rates derived by square-root batch-size scaling to $72$ for NOCS
and $96$ for HouseCat6D.
Our main accuracy results are averaged over three independent runs evaluated at the end of training.

\subsection{Comparison with prior methods}
\label{sec:comparison}
\label{sec:coverage}
\noindent\textbf{NOCS benchmarks.}
Table~\ref{tab:nocs} compares \ours{} with published joint detection and pose estimation results;
Table~\ref{tab:coverage} separately compares it with per-object pipelines under all-object evaluation.
FPS values in both tables are based on our RTX~A6000 measurements, with estimates marked in the captions.
In Table~\ref{tab:nocs}, we report IoU-matched pose AP for \ours{}; published joint results retain
their original evaluation populations.
CAD/prior counts instance-model supervision, augmentation, or explicit shape priors
in the reported pipelines, excluding synthetic-image generation and pose/coordinate annotations alone.
The accuracy comparison covers RGB-D methods;
YOPO~\cite{yopo} addresses the distinct RGB-only setting.
\ours{} achieves $85.7$ and $52.5$
$5^\circ5$\,cm AP on CAMERA25 and REAL275, respectively, exceeding the
strongest listed joint method, ShAPO, by $19.1$ and $3.7$ points.
\ours{} also achieves $98.6$ and $93.0$ IoU$_{50}$ on CAMERA25 and REAL275,
respectively, the highest values among the listed joint methods.
These comparisons include ShAPO's test-time refinement; its additional shape
and appearance reconstruction are outside the pose task evaluated here.

\noindent\textbf{All-object comparison with per-object pipelines.}
Table~\ref{tab:coverage} evaluates released per-object models~\cite{dpdn,hspose,vinet,agpose,cleanpose}
with the benchmark's supplied detections and masks, while \ours{} predicts its own.
All methods are scored on the same annotated frames using CleanPose's evaluator.
On CAMERA25, \ours{} achieves $85.8$ and $92.5$ AP at $5^\circ5$\,cm and $10^\circ5$\,cm,
exceeding CleanPose by $5.9$ and $4.4$ points, respectively, with the highest IoU$_{50}$
and CATRE IoU$_{75}$ among the evaluated models.
On REAL275, \ours{} reaches
$51.1$ at $5^\circ5$\,cm, within $1.6$ points of AG-Pose and $4.0$ of CleanPose,
while achieving the highest $10^\circ5$\,cm AP among the compared methods, at $73.6$.
AG-Pose and CleanPose remain stronger on REAL275 at tighter pose thresholds and CATRE IoU$_{75}$.
Re-scoring the same REAL275 predictions with IoU matching shows a $5^\circ5$\,cm AP
reduction of $1.4$ points for \ours{} under all-object evaluation,
compared with $8.5$--$12.3$ for the per-object pipelines.

\input{tables/nocs_all_objects}

\noindent\textbf{HouseCat6D.}
Table~\ref{tab:housecat} evaluates AG-Pose and CleanPose using the same
detections and masks as \ours{}. All annotated objects are scored, including those rejected by a
method's crop or depth-validity checks. Without a per-object crop stage,
\ours{} reaches $21.5$ AP at $5^\circ5$\,cm, comparable to CleanPose's $21.1$
and above AG-Pose's $17.7$. Both methods remain stronger in IoU$_{50}$ and
$10^\circ5$\,cm AP.
Common predicted inputs prevent differences between instance segmentors from
confounding the pose comparison. The GT-mask results provide a separate
reference for performance when accurate object regions are supplied.
The upper GT-mask block reproduces published IoU-matched pose results, while
the lower GT-mask block reports our evaluation of released models over all
annotations. IoU AP uses all annotations in both blocks.

\input{tables/housecat6d}

\subsection{Ablation studies}
\label{sec:ablations}
\input{tables/architecture_ablation}
\input{tables/component_controls}
\input{figures/captions/figure3_qualitative}
Tables~\ref{tab:components} and~\ref{tab:mechanisms} assess the integrated geometry-and-refinement
design at architectural and within-component levels, using the $12$-epoch NOCS training settings and all-object pose AP.

\noindent\textbf{Architectural contributions.}
The base RGB-D predictor retains depth anchoring, geometry-valued attention and
a local-3D-statistics MLP, without \modulelabel{A}--\modulelabel{C} or a learned depth encoder.
\ours{} improves $5^\circ5$\,cm AP by $18.2$ points over this base.
Removing both \modulelabel{B} branches gives $37.6$ AP; removing the entire \modulelabel{C}
and its auxiliary keypoint supervision gives $44.2$.
The \emph{Pooled descriptors + MLP} variant retains \modulelabel{A} and
replaces \modulelabel{B}/\modulelabel{C} with pooling and direct MLP pose prediction.
Both the \modulelabel{A}+\modulelabel{B} and \modulelabel{A}+\modulelabel{C} variants
fall below the full model on all displayed metrics. Together, these comparisons
support combining geometry-aware query updates with subsequent pose refinement.

\noindent\textbf{Early fusion and refinement representations.}
Our EdgeCrafter-based early-fusion~\cite{mvxnet} predictors use a SECOND-style
sparse-voxel CNN~\cite{second} or PointNet++~\cite{pointnetpp} for depth encoding, followed by query decoding
and direct pose heads, without \modulelabel{A}--\modulelabel{C}.
Both retain geometry-valued attention and remove metric depth anchoring.
The added depth encoder and cross-modal fusion layers use half the batch-size-scaled base
learning rate in Sec.~\ref{sec:implementation}; PointNet++ also uses FP32 encoder self-attention.
Their $5^\circ5$\,cm AP scores, $34.0$ and $29.8$, are both lower than \ours{}'s.
The two sparse-voxel hybrids instead retain \modulelabel{A}--\modulelabel{C}, including
geometric sampling, support weighting, and auxiliary supervision.
They differ only in refinement inputs:
adding geometry to fused RGB--depth features improves IoU$_{75}$ by $12.4$ points.
Both hybrids additionally enable FP32 encoder self-attention and sampled-feature
gradients, whereas these are disabled in \ours{}.
The results favor explicit object-associated geometry over relying solely on
an additional depth encoder in the tested set predictors.

\noindent\textbf{Within-component controls.}
In Table~\ref{tab:mechanisms}, mask weighting improves IoU$_{75}$ but trades off pose accuracy;
depth consensus gives higher values on all three displayed metrics.
Replacing the scene CNN with a pointwise MLP retains both feedback branches
but reduces $5^\circ5$\,cm AP to $36.9$.
With shared scene feedback retained, removing object feedback gives $51.3$
$5^\circ5$\,cm AP, compared with $51.1$ for \ours{}.
Removing both feedback branches instead gives $37.6$ AP (Table~\ref{tab:components}).
Together with the scene-MLP comparison, these results support spatially structured
scene context within the combined feedback pathway.

\noindent\textbf{Pose-state refinement.}
The \modulelabel{C} controls retain the remaining refinement components and auxiliary supervision.
Removing pose inputs slightly lowers $5^\circ5$\,cm AP
but improves IoU$_{75}$.
\emph{Static-pose conditioning} uses a fixed pose for later conditioning
while retaining recurrent updates.
\emph{Fixed-anchor prediction} instead applies averaged raw pose-head outputs to a
fixed anchor without predicted-pose feedback, changing the pose-output parameterization.
Their $5^\circ5$\,cm AP scores are $48.4$ and $44.5$, versus $51.1$ for \ours{}.
These controls distinguish providing a pose embedding from repeatedly coupling
subsequent predictions to an evolving pose state.
The larger reductions under static conditioning and fixed-anchor prediction indicate that
the pose-state update strategy matters beyond the presence of a pose input alone.
Taken together, Tables~\ref{tab:components} and~\ref{tab:mechanisms} support combining
object-associated geometry, shared scene context, and recurrent pose corrections,
with metric-dependent trade-offs among their internal operations.

\input{tables/runtime}
\subsection{Qualitative results}
\label{sec:qualitative}
Figure~\ref{fig:qualitative} compares \ours{} with AG-Pose and CleanPose on
REAL275 and HouseCat6D. The selected REAL275 examples show recovery of a laptop missed
by both compared methods, reduced laptop-height underestimation, and
camera position estimates closer to GT. On HouseCat6D, the bottle and tube illustrate
up-axis/extent and position differences, respectively; the symmetric bottle
does not establish canonical-yaw accuracy.
REAL275 uses released AG-Pose and CleanPose predictions; HouseCat6D shares \ours{}
detections and masks to compare pose estimates under common inputs.

The mug views reveal both pose errors and inconsistency in AG-Pose and CleanPose
predictions. The predicted orientations change substantially even between neighboring
evaluated frames ($t$, $t+3$), and larger deviations from GT reappear at
$t+11$, within a short 11-frame span. In contrast, \ours{} retains more accurate,
consistent orientations across these views, although each frame is processed
independently without tracking or temporal smoothing.

\subsection{Runtime and object-count scaling}
\label{sec:latency}
\label{sec:ryopon}
\label{sec:frontier}
\ours{} achieves $31.8$ FPS for full-frame pose estimation, including scene encoding
and object refinement, without an external detector.
Table~\ref{tab:latency} shows that per-object pipeline costs increase with object count.
With fixed input resolution and query capacity, \ours{} processes objects jointly
without repeated crop encoding.

We report mean timings on an RTX~A6000 at batch size $1$ after $50$ warm-up iterations,
from decoded inputs to host-side poses, excluding disk decoding and mask delivery.
We use COCO-pretrained Mask R-CNN (ResNet-50-FPN) to estimate detector time $D$;
accuracy evaluation retains the benchmark's supplied masks.
Per-object methods use eager FP32, with total times estimated as $D+P(K)$;
\ours{} is timed directly with TensorRT/PyTorch.
Deployment changes accuracy by at most $0.14$ pose AP points.

%% file: tables/nocs_joint.tex
\begin{table*}[tp]
\vspace*{6pt}
\caption{Joint RGB-D detection and pose estimation on NOCS. CAD/prior: instance 3D models or explicit shape priors used during training or inference; TTO: test-time optimization. $^\ddagger$short-run TTO FPS estimate.}
\label{tab:nocs}
\label{tab:nocse2e}
\centering\footnotesize\setlength{\tabcolsep}{5pt}
\begin{tabular}{@{}lcc@{\hskip 8pt}rrr@{\hskip 12pt}rrr@{\hskip 8pt}r@{}}
\toprule
 & & & \multicolumn{3}{c}{\textbf{CAMERA25}} & \multicolumn{3}{c}{\textbf{REAL275}} & \\
\cmidrule(lr){4-6}\cmidrule(lr){7-9}
Method & CAD/prior & TTO
 & IoU$_{50}$ & $5^\circ$5\,cm & $10^\circ$5\,cm
 & IoU$_{50}$ & $5^\circ$5\,cm & $10^\circ$5\,cm & FPS\\
\midrule
CenterSnap~\cite{centersnap} & \ding{51} & \ding{55} & 92.3 & 63.0 & 79.5 & \underline{80.2} & 27.2 & 58.8 & 28.9 \\
CenterSnap-R~\cite{centersnap} & \ding{51} & \ding{51} & 92.5 & 66.2 & 81.3 & \underline{80.2} & 29.1 & 64.3 & -- \\
ShAPO~\cite{shapo} & \ding{51} & \ding{51} & \underline{93.5} & \underline{66.6} & \underline{81.9} & 79.0 & \underline{48.8} & \underline{66.8} & 0.03$^\ddagger$ \\
FSD~\cite{fsd} & \ding{51} & \ding{55} & -- & -- & -- & 77.4 & 28.1 & 61.5 & -- \\
SceneGrasp~\cite{scenegrasp} & \ding{51} & \ding{55} & -- & -- & -- & 76.2 & 18.4 & 37.5 & -- \\
\oursrow{\ours{}} & \ding{55} & \ding{55} & \textbf{98.6} & \textbf{85.7} & \textbf{92.5} & \textbf{93.0} & \textbf{52.5} & \textbf{76.8} & 31.8 \\
\bottomrule
\end{tabular}
\vspace*{-6pt}
\end{table*}

%% file: tables/nocs_all_objects.tex
\begin{table*}[tp]
\vspace*{6pt}
\caption{NOCS all-object comparison with per-object pose pipelines. Pose AP includes all annotated instances. $^{*}$CATRE IoU; $^\dagger$detector-plus-pose FPS estimate at $K=8$.}
\label{tab:coverage}
\centering\footnotesize\setlength{\tabcolsep}{2.3pt}
\begin{tabular}{@{}l@{\hskip 5pt}rrrrrr@{\hskip 7pt}rrrrrr@{\hskip 5pt}r@{}}
\toprule
 & \multicolumn{6}{c}{\textbf{CAMERA25}} & \multicolumn{6}{c}{\textbf{REAL275}} & \\
\cmidrule(lr){2-7}\cmidrule(lr){8-13}
Method
 & IoU$_{50}$ & IoU$_{75}^{*}$ & $5^\circ$2\,cm & $5^\circ$5\,cm & $10^\circ$2\,cm & $10^\circ$5\,cm
 & IoU$_{50}$ & IoU$_{75}^{*}$ & $5^\circ$2\,cm & $5^\circ$5\,cm & $10^\circ$2\,cm & $10^\circ$5\,cm & FPS\\
\midrule
DPDN~\cite{dpdn} & 93.9 & 70.5 & 64.1 & 70.7 & 75.3 & 84.8 & 83.4 & 54.4 & 38.1 & 42.2 & 58.7 & 65.3 & 8.8$^\dagger$\\
HS-Pose~\cite{hspose} & 82.1 & 51.3 & 62.2 & 68.9 & 67.4 & 75.9 & 82.8 & 41.3 & 38.5 & 46.6 & 57.3 & 70.1 & 10.4$^\dagger$\\
VI-Net~\cite{vinet} & 91.4 & 77.2 & 70.2 & 77.0 & 74.5 & 82.0 & 83.1 & 48.4 & 40.7 & 47.1 & 58.7 & 68.3 & 8.7$^\dagger$\\
AG-Pose (DINO)~\cite{agpose} & \underline{94.1} & \underline{80.1} & 75.6 & 79.5 & 82.7 & 87.9 & \underline{84.1} & \underline{61.2} & \underline{46.4} & \underline{52.7} & \underline{63.2} & 71.0 & 7.5$^\dagger$\\
CleanPose~\cite{cleanpose} & \underline{94.1} & 79.6 & \underline{76.5} & \underline{80.0} & \underline{83.4} & \underline{88.1} & \underline{84.1} & \textbf{62.6} & \textbf{50.4} & \textbf{55.1} & \textbf{65.7} & \underline{72.3} & 7.5$^\dagger$\\
\oursrow{\ours{}} & \textbf{98.6} & \textbf{81.5} & \textbf{79.8} & \textbf{85.8} & \textbf{86.0} & \textbf{92.5} & \textbf{93.0} & 52.7 & 38.7 & 51.1 & 57.7 & \textbf{73.6} & 31.8\\
\bottomrule
\end{tabular}
\vspace*{-6pt}
\end{table*}

%% file: tables/housecat6d.tex
\begin{table}[tp]
\vspace*{6pt}
\caption{HouseCat6D. Shaded rows and GT denote ground-truth masks.}
\label{tab:housecat}
\definecolor{gtmask}{gray}{0.93}
\newcommand{\gtcell}[1]{\textcolor{black!55}{#1}}
\centering
\setlength{\tabcolsep}{1.8pt}\fontsize{7.2}{8.4}\selectfont
\begin{tabular}{ll@{\hskip 5pt}rrrrrr}
\toprule
Method & masks & IoU$_{25}$ & IoU$_{50}$ & $5^\circ$2\,cm & $5^\circ$5\,cm & $10^\circ$2\,cm & $10^\circ$5\,cm \\
\midrule
\multicolumn{8}{l}{\textbf{Pose AP: IoU-matched instances}} \\
\rowcolor{gtmask} \gtcell{VI-Net~\cite{vinet}} & \gtcell{GT} & \gtcell{80.7} & \gtcell{56.4} & \gtcell{8.4} & \gtcell{10.3} & \gtcell{20.5} & \gtcell{29.1} \\
\rowcolor{gtmask} \gtcell{AG-Pose (DINO)~\cite{agpose}} & \gtcell{GT} & \gtcell{88.1} & \gtcell{76.9} & \gtcell{21.3} & \gtcell{22.1} & \gtcell{51.3} & \gtcell{54.3} \\
\rowcolor{gtmask} \gtcell{CleanPose~\cite{cleanpose}} & \gtcell{GT} & \gtcell{89.2} & \gtcell{79.8} & \gtcell{22.4} & \gtcell{24.1} & \gtcell{51.6} & \gtcell{56.5} \\
\midrule
\multicolumn{8}{l}{\textbf{Pose AP: all annotated instances}} \\
\rowcolor{gtmask} \gtcell{AG-Pose (DINO)~\cite{agpose}} & \gtcell{GT} & \gtcell{87.9} & \gtcell{76.5} & \gtcell{21.5} & \gtcell{22.9} & \gtcell{50.6} & \gtcell{54.3} \\
\rowcolor{gtmask} \gtcell{CleanPose~\cite{cleanpose}} & \gtcell{GT} & \gtcell{90.1} & \gtcell{79.1} & \gtcell{24.6} & \gtcell{25.9} & \gtcell{53.1} & \gtcell{56.6} \\
AG-Pose (DINO)~\cite{agpose} & ours & \underline{70.3} & \underline{64.7} & 16.5 & 17.7 & \underline{43.9} & \underline{48.1} \\
CleanPose~\cite{cleanpose} & ours & \textbf{70.9} & \textbf{66.2} & \textbf{19.6} & \underline{21.1} & \textbf{46.1} & \textbf{51.2} \\
\oursrow{\ours{}} & ours & 64.7 & 57.8 & \underline{19.1} & \textbf{21.5} & 39.0 & 43.2 \\
\bottomrule
\end{tabular}
\vspace*{-6pt}
\end{table}

%% file: tables/architecture_ablation.tex
\begin{table}[t]
\vspace*{6pt}
\caption{Component configurations and early-fusion alternatives on REAL275.
$^*$CATRE IoU.}
\label{tab:components}
\label{tab:depthencoder}
\label{tab:earlyref}
\centering\footnotesize
\setlength{\tabcolsep}{2pt}
\begin{tabularx}{\columnwidth}{@{}Xcccccc@{}}
\toprule
& & & & \multicolumn{3}{c}{AP (\%) $\uparrow$} \\
\cmidrule(lr){5-7}
Variant & \modulelabel{A} & \modulelabel{B} & \modulelabel{C} & $\mathrm{IoU}_{75}^{*}$ & $5^\circ$5\,cm & $10^\circ$5\,cm \\
\midrule
\multicolumn{7}{@{}l}{\textit{RGB-D component configurations}} \\
Base RGB-D predictor & -- & -- & -- & 27.8 & 32.9 & 69.0 \\
Pooled descriptors + MLP & $\checkmark$ & -- & -- & 29.0 & 36.2 & 65.4 \\
w/o pose refinement & $\checkmark$ & $\checkmark$ & -- & 23.0 & 44.2 & 63.3 \\
w/o query feedback & $\checkmark$ & -- & $\checkmark$ & 49.9 & 37.6 & 70.6 \\
\rowcolor{black!5}
\textbf{\ours{}} (geom.+RGB) & $\checkmark$ & $\checkmark$ & $\checkmark$ & \textbf{52.7} & \textbf{51.1} & \textbf{73.6} \\
\midrule
\multicolumn{7}{@{}l}{\textit{Early-fusion alternatives}} \\
Sparse-voxel CNN: direct pose & -- & -- & -- & 24.2 & 34.0 & 67.4 \\
PointNet++: direct pose & -- & -- & -- & 23.6 & 29.8 & 66.6 \\
Sparse-voxel: fused features & $\checkmark$ & $\checkmark$ & $\checkmark$ & 36.0 & 48.5 & 66.6 \\
Sparse-voxel: geom.+fused & $\checkmark$ & $\checkmark$ & $\checkmark$ & 48.4 & 49.9 & 68.3 \\
\bottomrule
\end{tabularx}
\vspace*{-6pt}
\end{table}

%% file: tables/component_controls.tex
\begin{table}[t]
\vspace*{6pt}
\caption{Within-component controls on REAL275 (AP, \%). $^*$CATRE IoU.}
\label{tab:mechanisms}
\centering\footnotesize
\setlength{\tabcolsep}{3pt}
\begin{tabularx}{\columnwidth}{@{}Xccc@{}}
\toprule
Variant & $\mathrm{IoU}_{75}^{*}$ & $5^\circ$5\,cm & $10^\circ$5\,cm \\
\midrule
\rowcolor{black!5}
\textbf{\ours{}} & 52.7 & 51.1 & 73.6 \\
\midrule
\multicolumn{4}{@{}l}{\modulelabel{A} Query-conditioned geometry} \\
w/o mask weighting in pooling & 48.2 & \textbf{53.2} & \textbf{74.9} \\
w/o depth consensus & 49.6 & 50.6 & 71.0 \\
\midrule
\multicolumn{4}{@{}l}{\modulelabel{B} Geometry feedback} \\
Scene CNN replaced by MLP & 52.8 & 36.9 & 72.1 \\
w/o object feedback & 52.0 & 51.3 & 72.3 \\
\midrule
\multicolumn{4}{@{}l}{\modulelabel{C} Object-centric pose refinement} \\
w/o pose-conditioning input & \textbf{54.5} & 50.2 & 71.7 \\
Static-pose conditioning & 50.3 & 48.4 & 69.4 \\
Fixed-anchor prediction & 44.5 & 44.5 & 68.2 \\
\bottomrule
\end{tabularx}
\vspace*{-6pt}
\end{table}

%% file: figures/captions/figure3_qualitative.tex
\begin{figure*}[t]
\vspace*{4pt}
\centering
\includegraphics[width=0.948\textwidth]{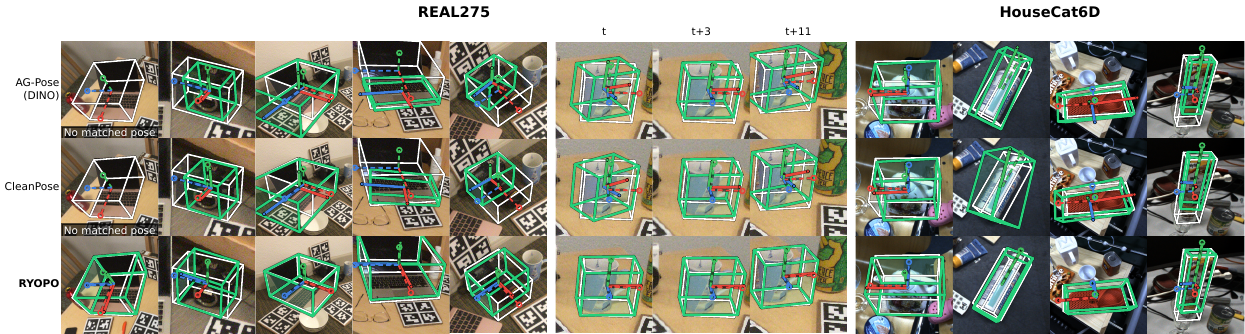}
\caption{Instance recovery, pose alignment, and nearby-frame consistency on REAL275 and HouseCat6D. Each column compares the same crop across methods. The mug views share a crop and illustrate pose accuracy and consistency across nearby observations; $t$, $t+3$, and $t+11$ denote source-frame offsets. All predictions are shown without yaw alignment at a $0.3$ confidence threshold. GT/predicted cuboids are white/green. Canonical axes are RGB: dashed with hollow tips for GT, solid with filled tips for predictions; symmetric objects show only the observable up axis.}
\label{fig:qualitative}
\vspace*{-4pt}
\end{figure*}

%% file: tables/runtime.tex
\begin{table}[!t]
\vspace*{6pt}
\caption{Pose-output latency and supplied-object scaling (ms/frame). $K$: supplied candidates; $P(K)$: cropped pose pipeline; \mbox{$D$: Mask R-CNN}.}
\label{tab:latency}
\centering\footnotesize
\setlength{\tabcolsep}{2pt}
\begin{tabular*}{\columnwidth}{@{\extracolsep{\fill}}lrrrrr@{}}
\toprule
& & \multicolumn{3}{c}{Cropped pipeline $P(K)$} & Total \\
\cmidrule(lr){3-5}
Method & $D$ & $K{=}1$ & $K{=}4$ & $K{=}8$ & $K{=}8$ \\
\midrule
CleanPose & 26.1 & 72.7 & 79.7 & 106.5 & 132.7 \\
AG-Pose & 26.1 & 71.0 & 82.9 & 107.3 & 133.4 \\
HS-Pose & 26.1 & 20.1 & 43.8 & 70.3 & 96.5 \\
VI-Net & 26.1 & 76.8 & 83.2 & 88.4 & 114.6 \\
DPDN & 26.1 & 54.8 & 69.0 & 88.1 & 114.2 \\
\midrule
\oursrow{\ours{}} & -- & \multicolumn{3}{c}{Joint full-frame inference} & \textbf{31.4} \\
\bottomrule
\end{tabular*}
\vspace*{-6pt}
\end{table}

%% file: sections/conclusion.tex
\section{Conclusion and Limitations}
\ours{} unifies RGB-D detection, segmentation, and \mbox{9-DoF} object pose estimation
in an end-to-end query-based set predictor, without explicit CAD-derived shape priors
or an external instance segmentor.
Its query-conditioned geometry pathway combines object observations and shared scene context
with refinement of an explicit pose state.
\ours{} surpasses published RGB-D joint results on NOCS and achieves competitive pose performance
compared with two-stage methods on REAL275 and HouseCat6D under all-object evaluation,
with $31.8$ FPS for full-frame pose output.
Pose accuracy at the tighter $2$\,cm translation threshold remains a limitation:
on REAL275, \ours{} achieves
lower $5^\circ2$\,cm and $10^\circ2$\,cm pose AP than AG-Pose and CleanPose.
Per-object methods also remain stronger in cuboid alignment and $10^\circ5$\,cm
pose AP on HouseCat6D.

%% file: sections/appendices.tex
\renewcommand{\thesection}{\arabic{section}}
\renewcommand{\thesectiondis}{\thesection.}
\renewcommand{\thesubsection}{\thesection.\arabic{subsection}}
\renewcommand{\thesubsectiondis}{\thesubsection.}
\renewcommand{\theHsection}{supplement.\arabic{section}}
\renewcommand{\theHsubsection}{\theHsection.\arabic{subsection}}
\setcounter{table}{0}
\setcounter{figure}{0}
\renewcommand{\thetable}{S\arabic{table}}
\renewcommand{\thefigure}{S\arabic{figure}}
\renewcommand{\theHtable}{supplement.\arabic{table}}
\renewcommand{\theHfigure}{supplement.\arabic{figure}}
\setlength{\textfloatsep}{9pt plus 2pt minus 1pt}
\setlength{\floatsep}{8pt plus 2pt minus 1pt}
\setlength{\intextsep}{7pt plus 1pt minus 1pt}

\makeatletter
\renewcommand{\section}{\@startsection{section}{1}{\z@}%
  {-2ex plus -0.5ex minus -0.2ex}{0.8ex}%
  {\normalfont\large\bfseries\raggedright}}
\renewcommand{\subsection}{\@startsection{subsection}{2}{\z@}%
  {-1.5ex plus -0.5ex minus -0.2ex}{0.7ex}%
  {\normalfont\normalsize\bfseries\raggedright}}
\newenvironment{appendixwidefloat}[1]{%
  \begin{minipage}{\textwidth}\def\@captype{#1}%
}{\end{minipage}}
\makeatother

\twocolumn[{
{\raggedright\normalfont\LARGE\bfseries Appendices\par}
\vspace{4pt}
\section{Mask-free Object Pose Estimation}
\label{sec:extended}
\label{sec:maskfree}
\input{tables/maskfree_results}
\vspace{8pt}
}]
\input{figures/captions/figureS1_maskfree}
\input{figures/captions/figureS2_maskfree_comparison}
\noindent\textbf{Variant and protocol.}
The mask-free variant removes the segmentation head, segmentation losses, and GT-mask
weighting in auxiliary correspondence supervision.
Box-aligned depth sampling uses $m_i=1$, so initial support becomes
$b_i=v_i$ in module~\modulelabel{A} (Fig.~\ref{fig:maskfree_geometry}).
Depth-consensus pooling, scene features, and pose refinement are retained.

Table~\ref{tab:arxiv_maskfree} compares main-table RYOPO averages with one
independently trained mask-free model per benchmark under all-object
evaluation (Sec.~\ref{sec:protocol}). The optimizer, learning-rate schedule,
sampling, and augmentation are unchanged.

\noindent\textbf{Accuracy trade-offs.}
Mask-free prediction is close to \ours{} at $5$\,cm on NOCS, with larger
gaps at $2$\,cm on CAMERA25. On HouseCat6D, it raises $5^\circ5$\,cm AP
from $21.5$ to $24.4$ but lowers IoU$_{75}$, $10^\circ2$\,cm AP, and
$10^\circ5$\,cm AP. Figure~\ref{fig:maskfree_qualitative} shows same-frame
pose comparisons.
The deployed mask-free model achieves $48.0\%$ all-object $5^\circ5$\,cm AP
on REAL275, compared with $48.5\%$ before deployment.

The mask-free variant removes the need for pixel-level mask supervision.
This reduced labeling requirement, together with higher inference FPS,
can lower the barrier to deploying category-level pose estimation in
real-world applications.

\section{Classification and Pose Trade-offs}
\label{sec:classification_weight}
We examine how classification-loss weighting affects detection and pose
accuracy in the mask-free model. Table~\ref{tab:arxiv_classification_weight}
varies this weight on HouseCat6D while keeping IoU-weighted focal targets,
class matching cost $2$, and motion-blur/photometric augmentation fixed.
For each setting, we train one model for $18$ epochs and evaluate it at the
end of training. We report 2D box AP$_{50}$, with 3D IoU and pose AP following
the all-object protocol in Sec.~\ref{sec:protocol}.

\input{tables/classification_weights}

Relative to weight $1$, weight $5$ raises box AP$_{50}$ from $86.9$ to $90.8$
but lowers $5^\circ5$\,cm AP from $21.9$ to $19.1$. Weight $0.5$ instead
improves pose AP while reducing box AP$_{50}$.
Higher detection AP does not necessarily improve pose AP.

Future work should investigate joint matching costs and learning objectives
that better balance detection quality and pose accuracy during query
assignment and optimization.

\FloatBarrier
\twocolumn[{
\section{Real-world Demonstrations}
\label{sec:real_demos}
\input{figures/captions/figureS2_hexbolt}
\vspace{4pt}
}]
We present qualitative mask-free RGB-D demonstrations on hex bolts and
toys, without ground-truth poses.

\noindent\textbf{Training-data annotation.}
For both demonstrations, five ArUco markers placed in the scene provided
camera-pose estimates for the training captures. Within each sequence,
the objects remained stationary while only the camera moved.
We selected a representative frame and manually annotated each object's
pose and metric dimensions. We then propagated the annotations to the
remaining frames using the relative camera poses, transforming object
rotation and translation into each frame's camera coordinates while
keeping its dimensions unchanged.

\noindent\textbf{Training settings.}
Both models are initialized from pretrained RGB backbones and trained
separately on their captured datasets for $48$ epochs. We use binary
classification targets, classification-loss weight $5$, class matching cost
$4$, and motion-blur/photometric augmentation.
The base learning rate is $2\!\times\!10^{-4}$ at batch size $24$, with
square-root batch-size scaling and $0.1$-factor decays at epochs $32$ and $44$.
Bolt and toy batch sizes are $24$ and $48$, respectively.

\subsection{Live hex-bolt demonstration}
\label{sec:hexbolt}
\label{sec:hexbolt_live}
We demonstrate multi-object prediction with a RealSense L515 and RTX~4080
Super (Fig.~\ref{fig:hexbolt_live}).

Training uses $2{,}719$ RGB-D frames from $17$ captures, with
segmentation-assisted tight boxes. For labeling convenience, we treat
rotation about each bolt's local Y axis as a continuous symmetry, leaving
axial yaw unspecified rather than resolving the hexagonal head's discrete
orientations. The recording shows successive bolt placement, hand
occlusion, and viewpoint changes, with up to $16$ displayed detections.

Detections remain visually consistent across successive frames despite
pronounced motion blur and rapid camera motion, providing qualitative
evidence of detection robustness under these conditions.

\noindent\textbf{Latency and streaming rate.}
Online inference takes approximately $18$\,ms from
BGR-to-RGB conversion to CPU-side predictions, including preprocessing,
transfers, scene encoding, TensorRT inference, and postprocessing with
thresholding. This corresponds to about $56$ FPS of processing capacity,
while the camera limits the delivered rate to $30$ FPS.
Engine-level inference takes $10.3$\,ms ($97.1$ FPS) on preprocessed GPU
inputs, including scene encoding and postprocessing, with CUDA synchronization.
Both timings exclude camera acquisition, alignment/rectification, temporal
stabilization, rendering, and streaming.

\input{figures/captions/figureS3_toys}

\subsection{Qualitative transfer to unseen toys}
\label{sec:plush_live}
The toy model uses all $3{,}456$ RGB-D frames from $12$ videos of one physical
plush toy, with no rotational symmetry.
In addition to the shared training settings, the toy model uses foreground
RGB-D reprojection with probability $0.5$
(camera roll $\pm90^\circ$, object-local yaw $\pm20^\circ$ and pitch
$\pm10^\circ$), and depth dropout with probability $0.35$, removing
$20$--$60\%$ of valid foreground depth. Segmentation masks are used only
for augmentation and box construction, not as model inputs or supervision.

Figure~\ref{fig:plush_live} shows the training instance and two unseen
instances, another plush toy and a plastic toy, without additional training.
Despite differences in proportions, appearance, and material, the predicted
cuboids and axes remain visually aligned with the objects across the shown
viewpoints. These examples suggest cross-instance pose generalization to
similarly shaped objects from a limited capture set of one physical instance.

%% file: tables/maskfree_results.tex
\begin{appendixwidefloat}{table}
\centering
\caption{RYOPO and its mask-free variant. All-object AP (\%); $^{*}$CATRE IoU, other IoU independent; $^{\dagger}$deployed FPS on REAL275.}
\label{tab:arxiv_maskfree}
\footnotesize\setlength{\tabcolsep}{6pt}
\begin{tabular}{@{}lrrrrrrrr@{}}
\toprule
Variant & IoU$_{50}$ & IoU$_{75}$ & IoU$_{75}^{*}$ & $5^\circ$2\,cm & $5^\circ$5\,cm & $10^\circ$2\,cm & $10^\circ$5\,cm & FPS$^{\dagger}$\\
\midrule
\multicolumn{9}{c}{\textbf{CAMERA25}}\\
\ours{} & \textbf{98.6} & \textbf{96.7} & \textbf{81.5} & \textbf{79.8} & \textbf{85.8} & \textbf{86.0} & \textbf{92.5} & -- \\
Mask-free & 98.4 & 96.3 & 78.5 & 77.0 & 85.1 & 83.4 & 92.0 & -- \\
\midrule
\multicolumn{9}{c}{\textbf{REAL275}}\\
\ours{} & \textbf{93.0} & \textbf{84.5} & \textbf{52.7} & \textbf{38.7} & \textbf{51.1} & \textbf{57.7} & \textbf{73.6} & 31.8 \\
Mask-free & 91.9 & 81.3 & 52.0 & 36.7 & 48.5 & 56.8 & 71.9 & \textbf{51.4} \\
\midrule
\multicolumn{9}{c}{\textbf{HouseCat6D}}\\
\ours{} & 57.8 & \textbf{34.4} & \textbf{30.5} & 19.1 & 21.5 & \textbf{39.0} & \textbf{43.2} & -- \\
Mask-free & \textbf{60.2} & 28.2 & 25.7 & \textbf{19.3} & \textbf{24.4} & 33.5 & 41.1 & -- \\
\bottomrule
\end{tabular}
\end{appendixwidefloat}

%% file: figures/captions/figureS1_maskfree.tex
\begin{figure}[t]
\vspace*{4pt}
\centering
\raisebox{-0.0pt}[138.0461pt][0.0pt]{%
  \makebox[245.71875pt][l]{\includegraphics{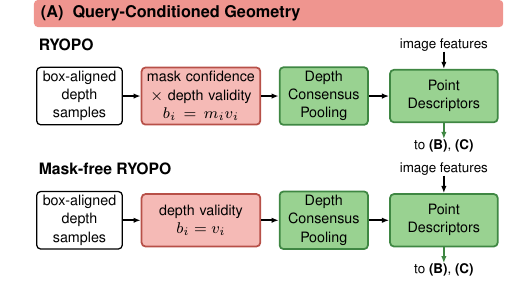}}}
\caption{Mask-free query-conditioned geometry. Box-aligned samples use
depth validity $v_i$ without mask confidence $m_i$. Depth-consensus pooling,
descriptor construction, geometry feedback \modulelabel{B}, and
pose refinement \modulelabel{C} are retained.}
\label{fig:maskfree_geometry}
\vspace*{-4pt}
\end{figure}

%% file: figures/captions/figureS2_maskfree_comparison.tex
\begin{figure}[!b]
\centering
\includegraphics[width=0.90\columnwidth]{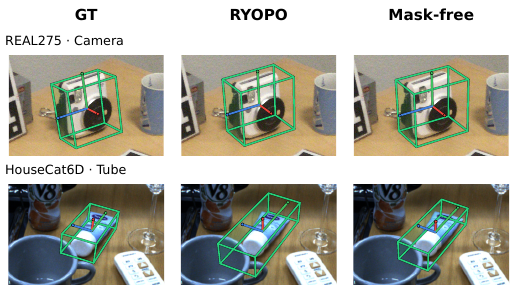}
\caption{RYOPO and mask-free pose predictions on the same REAL275 and
HouseCat6D object crops. Green cuboids show object extent; red, green,
and blue axes indicate the canonical X, Y, and Z directions.}
\label{fig:maskfree_qualitative}
\end{figure}

%% file: tables/classification_weights.tex
\begin{table}[!ht]
\centering
\caption{Classification-loss weight on HouseCat6D with mask-free \ours{}.
Scores in \%; $^{*}$CATRE IoU; pose AP uses all objects.}
\label{tab:arxiv_classification_weight}
\footnotesize\setlength{\tabcolsep}{4pt}
\begin{tabular}{@{}ccccc@{}}
\toprule
Cls. weight & Box AP$_{50}$ & IoU$_{75}^{*}$ & $5^\circ$5\,cm & $10^\circ$5\,cm\\
\midrule
0.5 & 83.1 & \underline{24.2} & \textbf{24.6} & \textbf{42.6}\\
1.0 & \underline{86.9} & \textbf{26.9} & \underline{21.9} & \underline{42.2}\\
2.0 & 85.2 & 20.8 & 21.8 & 40.2\\
3.0 & 85.8 & 20.1 & 14.6 & 38.8\\
5.0 & \textbf{90.8} & 20.8 & 19.1 & 39.4\\
\bottomrule
\end{tabular}
\end{table}

%% file: figures/captions/figureS2_hexbolt.tex
\begin{appendixwidefloat}{figure}
\vspace*{4pt}
\centering
\setlength{\abovecaptionskip}{0pt}
\includegraphics{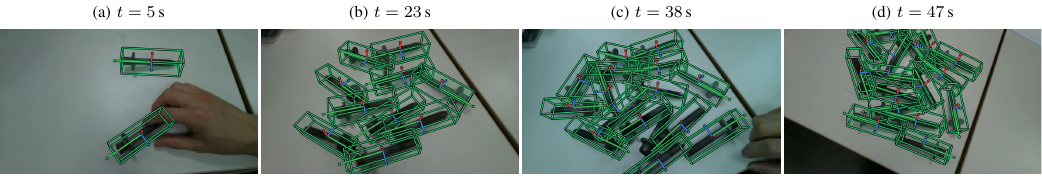}
\caption{Live mask-free hex-bolt predictions during object placement,
partial hand occlusion, and camera viewpoint changes.
Green cuboids and colored axes are predictions.
Times refer to the source recording.}
\label{fig:hexbolt_live}
\vspace*{-4pt}
\end{appendixwidefloat}

%% file: figures/captions/figureS3_toys.tex
\begin{figure}[t]
\centering
\setlength{\abovecaptionskip}{0pt}
\includegraphics{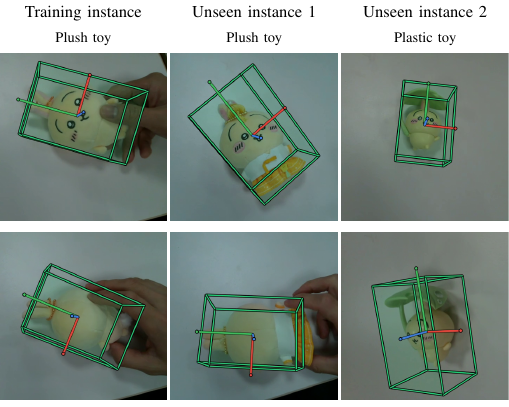}
\caption{Live predictions on the training toy and two unseen instances.
Top: front-facing views. Bottom: rear views (left, center) and an oblique
side view (right). The training instance is partially hand-occluded.}
\label{fig:plush_live}
\end{figure}